\documentclass[10pt,twocolumn,letterpaper]{article}

 \usepackage[pagenumbers]{iccv} 
\usepackage{multirow}
\usepackage{graphics}
\usepackage{bbding}
\usepackage{pifont}

\definecolor{iccvblue}{rgb}{0.21,0.49,0.74}
\usepackage[pagebackref,breaklinks,colorlinks,allcolors=iccvblue]{hyperref}

\def\paperID{3052} 
\def\confName{ICCV}
\def\confYear{2025}

\title{EgoMe: A New Dataset and Challenge for Following Me via Egocentric View in Real World}

\author{Heqian Qiu, Zhaofeng Shi, Lanxiao Wang, Huiyu Xiong , Xiang Li , Hongliang Li\\
	University of Electronic Science and Technology of China\\
	Chengdu, China\\
	{\tt\small hqqiu@uestc.edu.cn, zfshi@std.uestc.edu.cn,lanxiaowang@uestc.edu.cn,}\\
	{\tt\small hyxiong@std.uestc.edu.cn,202411012315@std.uestc.edu.cn,hlli@uestc.edu.cn}
}

\begin{document}
\maketitle
\begin{abstract} 

In human imitation learning, the imitator typically take the egocentric view as a benchmark, naturally transferring behaviors observed from an exocentric view to their owns, which provides inspiration for researching how robots can more effectively imitate human behavior. However, current research primarily focuses on the basic alignment issues of ego-exo data from different cameras, rather than collecting data from the imitator's perspective, which is inconsistent with the high-level cognitive process. To advance this research, we introduce a novel large-scale egocentric dataset, called EgoMe, which towards following the process of human imitation learning via the imitator's egocentric view in the real world. Our dataset includes 7902 paired exo-ego videos (totaling15804 videos)  spanning diverse daily behaviors in various real-world scenarios. For each video pair, one video captures an exocentric view of the imitator observing the demonstrator's actions, while the other captures an egocentric view of the imitator subsequently following those actions. Notably, EgoMe uniquely incorporates exo-ego eye gaze, other multi-modal sensor IMU data and different-level annotations for assisting in establishing correlations between observing and imitating process. We further provide a suit of challenging benchmarks for fully leveraging this data resource and promoting the robot imitation learning research. Extensive analysis demonstrates significant advantages over existing datasets. Our EgoMe dataset and benchmarks are available at https://huggingface.co/datasets/HeqianQiu/EgoMe.


\end{abstract}
\section{Introduction}
In daily life, human beings can observe the behaviors of other individuals and follow them, naturally transferring the acquired knowledge and skills to their own. This process not only reflects the natural cognitive instinct of human imitating learning but also provides profound insights into the exploration of robotic learning. By endowing robots with the capability to imitate human behavior, it is significant to achieve rapid skill improvement and diversified skill accumulation, especially in complex tasks, which holds profound significance for the next generation of AI agents including embodied intelligence \cite{duan2022survey} and AR/VR \cite{liu2022joint, wang2023scene}. To effectively acquire this ability, it is essential for the robot to directly observe and follow human behaviors from its own viewpoint, encompassing its dual roles as both an observer (exocentric view), meticulously recording and analyzing actions of demonstrator, and as a imitator (egocentric view), actively following and imitating others' behaviors. 
\begin{figure}[!t]
	\centering
	\includegraphics[width=1.0\linewidth]{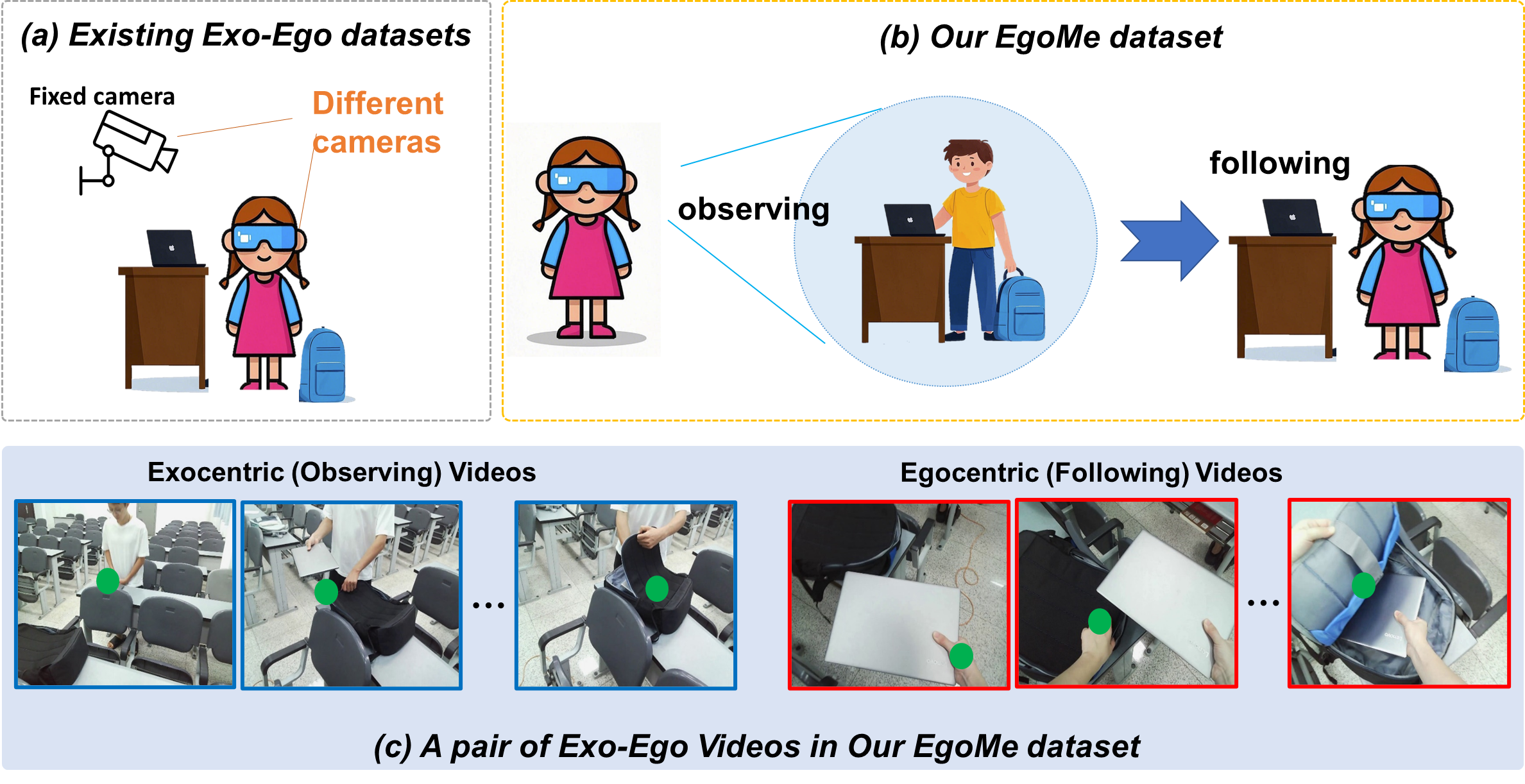}
	\caption{Our EgoMe dataset takes the egocentric view as a benchmark, following behaviors observed from a exocentric view to their own actions for exploring the process of human imitation learning in the real world.}
	\label{fig:introduction}
\end{figure}

In the early stage,  tremendous amount of research primarily focus on observing and analyzing human behavior via an exocentric (Exo) view \cite{kay2017kinetics, goyal2017something, caba2015activitynet, damen2018scaling, caba2015activitynet, kuehne2011hmdb, li2025zeroi2v, ashutosh2023hiervl, sun2022long, ye2023hitea, luiten2021hota, nagaraja2016modeling} (i.e., the observer's view). Although these methods can capture the external manifestations of behavior, it often struggles to deeply understand and appreciate the motivations and internal logic behind these behaviors. With the emergence of wearable devices, many research opportunities have arisen for a egocentric (Ego) view, which captures human perception and behavioral interactions in the real world.  However, robots often face complex ever-changing environments and tasks that cannot be fully described by a single view alone. 

Recently, a few researchers have made attempted to establish a connection between Exo and Ego views,  which can be roughly divided into two directions. One direction \cite{huang2024egoexolearn,xue2023learning} asynchronously integrates existing exocentric and egocentric datasets from different environments. Due to significant differences in dataset size, quality, and environmental contexts, it is limited in achieving precise alignment between the two views. Another direction \cite{grauman2024ego,li2024egoexo} simultaneously collects both ego-exo data from demonstrators using multiple cameras in the same environment. Although this way covers the consistency behavioral data from two different views as shown in Fig. \ref{fig:introduction}, both of them only record video information from the same demonstrator, making it difficult to reflect the complex and high-level cognitive process of how robots can simulate human observation and learning. This study will be an important breakthrough for the new development of next-generation AI, yet unfortunately, no efforts have been made thus far.

To take one more step forward, in this paper, we are the first to introduce a novel large-scale egocentric dataset named EgoMe, which aims to deeply capture and analyze the process of observing and following others' behaviors via egocentric view in the real world in Fig.\ref{fig:introduction}. In EgoMe, we record a pair of videos via the imitator's egocentric viewpoint wearing head-mounted glasses. One video captures an exocentric view of the imitator observing the demonstrator's actions, while the other captures an egocentric view of the imitator subsequently mimicking those actions. These two videos are recorded in the same scenario, targeting the same action, aiming to adequately demonstrate the human learning path from observation to imitation.

Specifically, our EgoMe dataset contains 15804 videos including 7902 pairs of videos with a total video duration of 82 hours and 46 minutes. It considers 184 activity categories from simple daily bodily movements to complex professional skill operations across 41 diverse scenarios in real-world environments, ensuring that the dataset fully reflects human learning behaviors that may be encountered in real life. Notably, in addition to video data, our dataset also contain eye gaze and other sensor data during the observing and following process. These additional pieces of information not only reflect the visual attention allocation and body posture behavioral characteristics of the wearer but also provides powerful data support for establishing correlations between observation and imitation behaviors.

Based on the collected data, we carefully label a series of annotations, encompassing eye gaze, action labels, coarse/fine-level video language annotations, and imitative verification annotations. Meanwhile, we perform a more comprehensive and in-depth statistical analysis for the EgoMe dataset, which demonstrates significant advantages compared to existing datasets. Accordingly, to fully leverage this data resource and propel research progress in related fields, we meticulously designed a suit of challenging benchmark tasks aimed at training and evaluating the learning capabilities of AI agent models through various application and testing scenarios. In addition, we pretrained a exo-ego cross-modal retrieval model to extract aligned exo-ego features for subsequent exo-ego gaze prediction, imitative action assessment/reorganization and coarse/fine-level video understanding. We hope that our dataset can aid in exploring more complex human learning process and provide new insights for the development of next-generation robots in the future.

%
%
%
%

\begin{table*}[!htbp]
	\caption{Comparison with other related egocentric datasets on settings (left) and annottaions (right). ``IMU" includes accelerometer, gyroscope and magnetometer. \ding{109}: partially contained, \ding{51}: contained, \ding{55}: without, \ding{111}: action label not language annotations.}
	\label{dataset_compar}
	\resizebox{\linewidth}{!}{\begin{tabular}{c|ccccccc|cccccccc}
			\toprule
			Dataset & Settings & Videos & Hours & Scenarios & Actions & Exo+Ego? &Imitator Perspective& Gaze & IMU& Coarse-level Language&Fine-level Language&Verification\\
			\midrule
			Meccano\cite{ragusa2021meccano} & Industry &20& 7 & 20 & 61 &  \ding{55}  &  \ding{55}   &   \ding{51}  &  \ding{55}  &  \ding{55} &  \ding{111}  &  \ding{55}    \\
			HoloAssist\cite{wang2023holoassist} & Assistive &350& 166 & - & 414 &  \ding{55}  &  \ding{55}   &   \ding{51}  &  \ding{51}  &  \ding{111} &  \ding{111}  &  \ding{51}    \\
			EGTEA\cite{li2018eye} & Cooking &28& 28 & - & 44 &  \ding{55}  &  \ding{55}   &   \ding{51}  &  \ding{55}  &  \ding{55} &  \ding{111}  &  \ding{55}    \\
			Epic-Kitchens-100\cite{damen2022rescaling} & Cooking &700& 100 & 45 & 4025 &  \ding{55}  &  \ding{55}   &   \ding{51}  &  \ding{109} &  \ding{55} &  \ding{111}  &  \ding{55}    \\
			Ego4D\cite{grauman2022ego4d} & Multiple &991& 3670 & 74 & 113 &  \ding{55}  &  \ding{55}   &   \ding{109}  &  \ding{109} &  \ding{111}  &  \ding{109}   &  \ding{55}    \\
			\midrule
			H2O\cite{kwon2021h2o} & Desk &500& 5 & 36 & 36 &  \ding{51}  &  \ding{55}   &   \ding{55}  &  \ding{55} &  \ding{55}  &  \ding{111}   &  \ding{55}    \\
			Assembly101\cite{sener2022assembly101} & Desk &4321& 513 &15&1380&  \ding{51}  &  \ding{55}   &   \ding{55}  &  \ding{111}  &  \ding{111}   &  \ding{55}    \\
			LEMMA\cite{jia2020lemma} & Daily &324& 10.1 &15&641&  \ding{51}  &  \ding{55}   &   \ding{55}  &  \ding{55} &  \ding{55}  &  \ding{111}   &  \ding{55} \\ 
			Homage\cite{rai2021home} & Daily &5700& 25.4 &70&-&\ding{51}  &  \ding{55}   &   \ding{55}  &  \ding{109} &  \ding{111}   & \ding{111}  &  \ding{55} \\ 
			Charades-Ego\cite{sigurdsson2018charades} & Daily &7860& 68.8 &15&157&\ding{51}  &  \ding{55}   &   \ding{55}  &  \ding{55} &  \ding{55}   & \ding{111}  &  \ding{55} \\ 
			EgoExoLearn~\cite{huang2024egoexolearn} & Daily \& Lab &747& 120 &8&39&\ding{51}  &  \ding{55}   &   \ding{109}  &  \ding{51} &  \ding{51}   & \ding{51}  &  \ding{55} \\ 
			Ego-Exo4D~\cite{grauman2024ego} & Multiple &5035& 1286.3 &43&689& \ding{51}  &  \ding{55}   &   \ding{109}  &  \ding{109} &  \ding{109}   & \ding{109}  &  \ding{55} \\ 
			EgoMe (ours) & Multiple & 15804 &82.8& 41& 184& \ding{51}  &  \ding{51}   &   \ding{51}  &  \ding{51} &  \ding{51}   & \ding{51}  &  \ding{51} \\ 
			\bottomrule
	\end{tabular}}
\end{table*}
\section{Related Works}

\noindent\textbf{Egocentric Datasets.} 
In recent years, egocentric datasets have become increasingly important for AR/VR and robotics. Some datasets primarily focus on unscripted daily life scenarios, including specific kitchen activities \cite{damen2018scaling, damen2022rescaling, tschernezki2024epic} with actions like food preparation, cooking, cutting and daily interactions with objects \cite{grauman2022ego4d,lee2012discovering,pirsiavash2012detecting,singh2016krishnacam,lv2024aria,ma2025nymeria,gu2025egolifter}. Among of them, Ego4D dataset \cite{grauman2022ego4d} provides an even broader range of environments, both indoor and outdoor, allowing for a more comprehensive study of human behavior in different settings. On the other hand, there are datasets that center around procedural tasks, which provides a certain activity instructions for demonstrators, such as EGTea~\cite{li2018eye}, AssistQ~\cite{wong2022assistq}, CaptainCook4D~\cite{peddi2023captaincook4d}, Meccano~\cite{ragusa2021meccano}, Epic-tent~\cite{jang2019epic}, and EgoProcel~\cite{bansal2022my}. These datasets are valuable for studying human actions when following a set of predefined procedures.


%


\noindent\textbf{Exo-Ego Datasets.}
Despite considerable efforts have been dedicated to egocentric research, it is crucial to build the interaction between different views in the real world for AI agents. Early datasets such as CMU-MMAC\cite{de2009guide} and Charades-Ego\cite{sigurdsson2018charades} provide paired egocentric with a camera fixed to their forehead and exocentric videos via a fixed camera, supporting simple action recognition in specific kitchen and household environments. Homage~\cite{rai2021home} offers hierarchical action labels for fine-grained activity recognition in household tasks. H2O\cite{kwon2021h2o} and Assembly101\cite{sener2022assembly101} provide detailed hand-object pose annotations and multi-step workflows at a lab tabletop. Lemma\cite{jia2020lemma} emphasizes multi-view learning using 3D skeleton and RGB-D data. The AE2 dataset\cite{xue2023learning}, by contrast, focuses on learning view-invariant embeddings for action recognition by using unpaired ego-exo videos from five different datasets. Recently, Ego-Exo4D~\cite{grauman2024ego} further delivers a large-scale multi-view and multi-modal data using different cameras for skilled human activities such as audio, eye gaze, 3D point clouds and so on. Additionally, EgoExoLearn\cite{huang2024egoexolearn} aims to bridge asynchronous egocentric videos and exocentric demonstration videos for procedural activities in daily-life assistance and professional support. EgoExo-Fitness\cite{li2024egoexo} focuses on specific full-body fitness activities recorded via  exocentric videos from three views and an egocentric video, emphasizing temporal and spatial consistency for action recognition. However, these datasets usually collect paired ego-exo videos from different perspectives without making an effort to explore human mimetic learning. In contrast, as shown in Table\ref{dataset_compar} our dataset records videos from the imitator's view and label multi-modal annotations to consistently simulate the interaction and learning between human and real-world for deeper cognitive understanding of embodied intelligence.
\begin{figure*}[!t]
	\centering
	\includegraphics[width=1.0\linewidth]{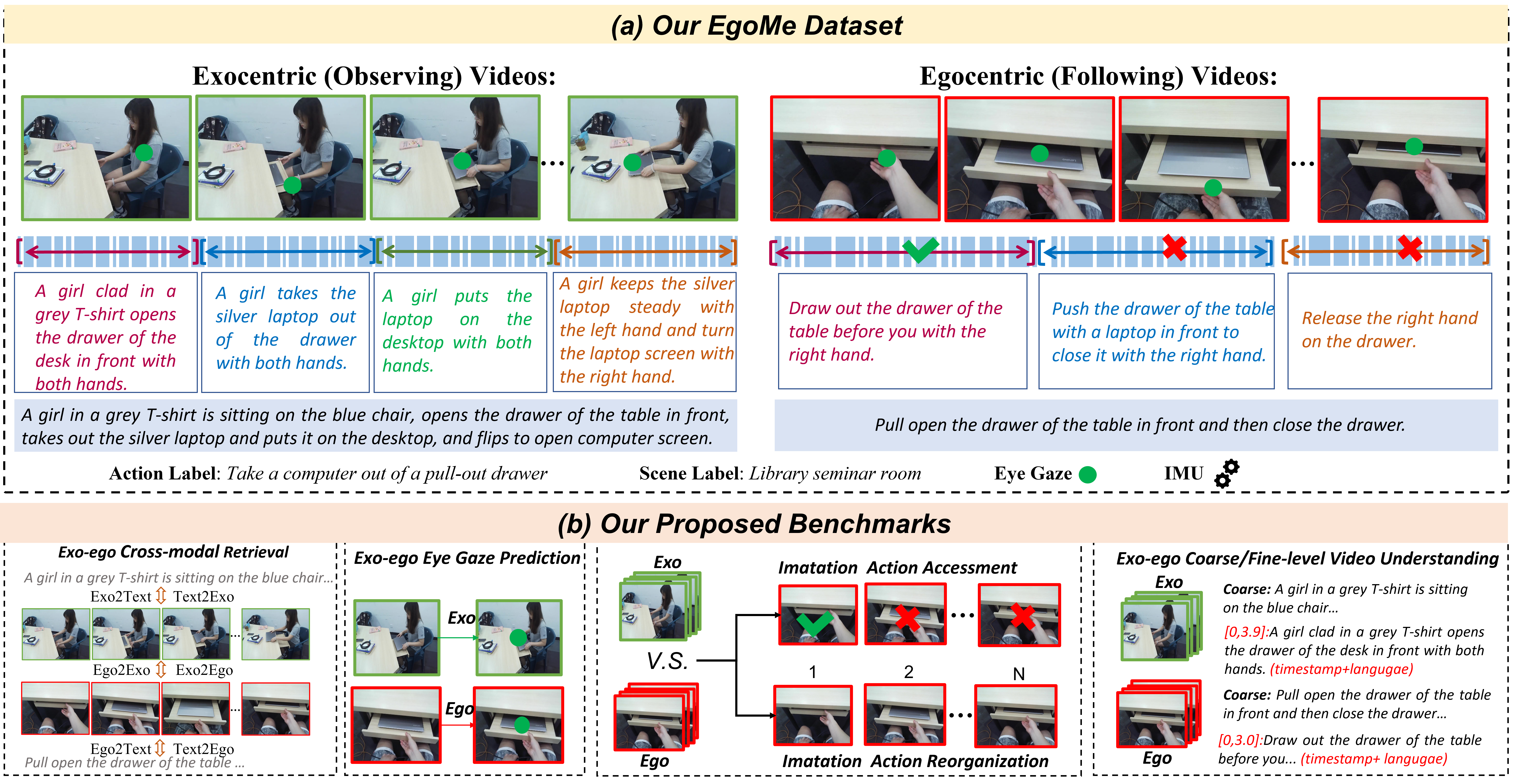}
	\caption{Illustrative example in our EgoMe dataset. It includes eye gaze, action label, coarse/fine-level language annotations, and imitative verification annotation during the observing (exocentric) and following (egocentric) process.}
	\label{fig:dataset}
\end{figure*}

\noindent\textbf {Video Understanding Benchmarks.} Nowadays, as the number of videos increases, video understanding has garnered the attention of a wide range of researchers. It encompasses a series of benchmarks, such as video classification or retrieval \cite{deng2024text,karpathy2014large,kay2017kinetics,yu2018joint}, action recognition or detection \cite{pirsiavash2012detecting,caba2015activitynet,kuehne2011hmdb}, video caption generation\cite{jang2017tgif, yu2018joint, torabi2016learning, sun2022long, ye2023hitea, chen2011collecting, xu2016msr}, procedure understanding \cite{grauman2022ego4d,sener2022assembly101,zhou2022survey, lee2024error} and so on. To bridge the gap between egocentric and exocentric videos, cross-view exo-ego related benchmarks has recently begun to emerge, which can be broadly categorized into several main types: exo-ego video alignment, which explores extracting features invariant to perspective changes or bridging asynchronous ego-exo data\cite{xue2023learning, huang2024egoexolearn, li2024egoexo}; exo-ego video generation that generates video sequences from different views; exo-ego hand-object interaction and object manipulation, targeting detailed hand-object dynamics in complex environments\cite{kwon2021h2o, sener2022assembly101,li2024egoexo}; multi-modal behavior assessment and proficiency evaluation, emphasizing key step recognition and skill assessment in multi-modal contexts\cite{grauman2024ego, kastingschafer2024seed4d}; cross-view action recognition and segmentation, focusing on identifying and analyzing actions in daily, household, or multi-step workflows\cite{de2009guide, sigurdsson2018charades, rai2021home, jia2020lemma, sener2022assembly101}. Unlike existing benchmarks, we propose a suite of new challenges for evaluating the whether imitators follow correctly observed behavior, such as exo-ego gaze prediction for attention process understanding, imitative action assessment and reorganization, and hierarchical procedural description of imitation behavior processes.

\section{Dataset}
To prompt the research on imitating learning, we introduce a new and challenge EgoMe dataset to mimic the human behavior in real world as shown in Fig.\ref{fig:dataset}. In this section, we first describe the pipeline of data collection, and various annotations, and then present the informative statistics and comparisons.

%

\subsection{Data Collection}
\label{sec:datacollection}
Since the data is expected to be collected via the imitator view, we adopt a head-mounted 7 invensun aSee Glasses as the data recording device, which can capture a 1280$\times$960 video with eye gaze and other sensor data like pupil information, saccade, blink, quaternion, angular velocity, acceleration, and magnetic strength. Unlike existing datasets with specific kitchen or labs scenarios, we consider a wider range of human daily activities across different real-world scenarios, enabling the robots assist humans more generally. In our EgoMe dataset, we collect data in up to over 41 scenarios including various classrooms, supermarkets, offices and so on. In addition, each kind of scenario comprises a diverse range of layouts.


For the data collection, we recruited 37 volunteers with their consent and followed strict privacy and ethical guidelines. Before the start of data collection, we require volunteers to carefully read and sign the data collection informed consent form and train them to operate the recording device and collection procedures. During the data collection, we set the volunteer group with at least 3 participants: one wearing the head-mounted glasses as an imitator, and the others as demonstrators. For the observing video, the imitator with glasses wearer is required to carefully observe the demonstrator performing the human activity via exocentric view. After observing, the imitator needs to follow and mimic the observed activity with the completely same actions, or is asked to conduct the activity with wrong action steps or actions in the egocentric view. Finally, we post-processed the recorded data and edited it into the egocentric (imitating) and exocentric (observing) data pairs.

\subsection{Annotations}
\label{sec:anno}
To support future studies on our EgoMe dataset, we provide five types of annotations in this section, among which we additionally recruit 30 proficient annotators to assist with large-scale language annotations and imitative verification annotations. The annotation quality is strictly controlled and checked by at least three individuals and checking code (including syntax checks using LLM models).
%

\noindent\textbf{Eye Gaze Annotation.} The eye gaze annotations are automatically recorded at a sampling rate of 120Hz along with the videos. To guarantee the quality of the eye gaze, the imitator is ask to complete the gaze calibration. The recorded eye gaze is binocular and contains the positions of left/right eye gaze points, vector, and the duration, position of eye fixation. For more precise annotations, the device also records pupil information including pupil validity, pupil diameter, saccade information such as duration, amplitude, and velocity, and eye blink information such as blink duration.


\noindent\textbf{Action Label Annotation.}
In our EgoMe dataset, we annotate video-level action labels with 184 common categories in 41 real-world daily scenarios, ranging from simple actions like typing on a phone to complex ones like folding clothes or using various instruments. In addition, considering the potential applications of scenes where AI robots assist or nurse humans, we innovatively introduce assisting and nursing-related actions, e.g. lifting objects with humans and helping humans wear clothes. Moreover, to ensure the dataset diversity, different videos within the same action category contain diverse subjects, actors, scenarios, camera angles, and contextual actions.
%
%

\begin{figure*}[!t]
	\centering
	\includegraphics[width=1.0\linewidth]{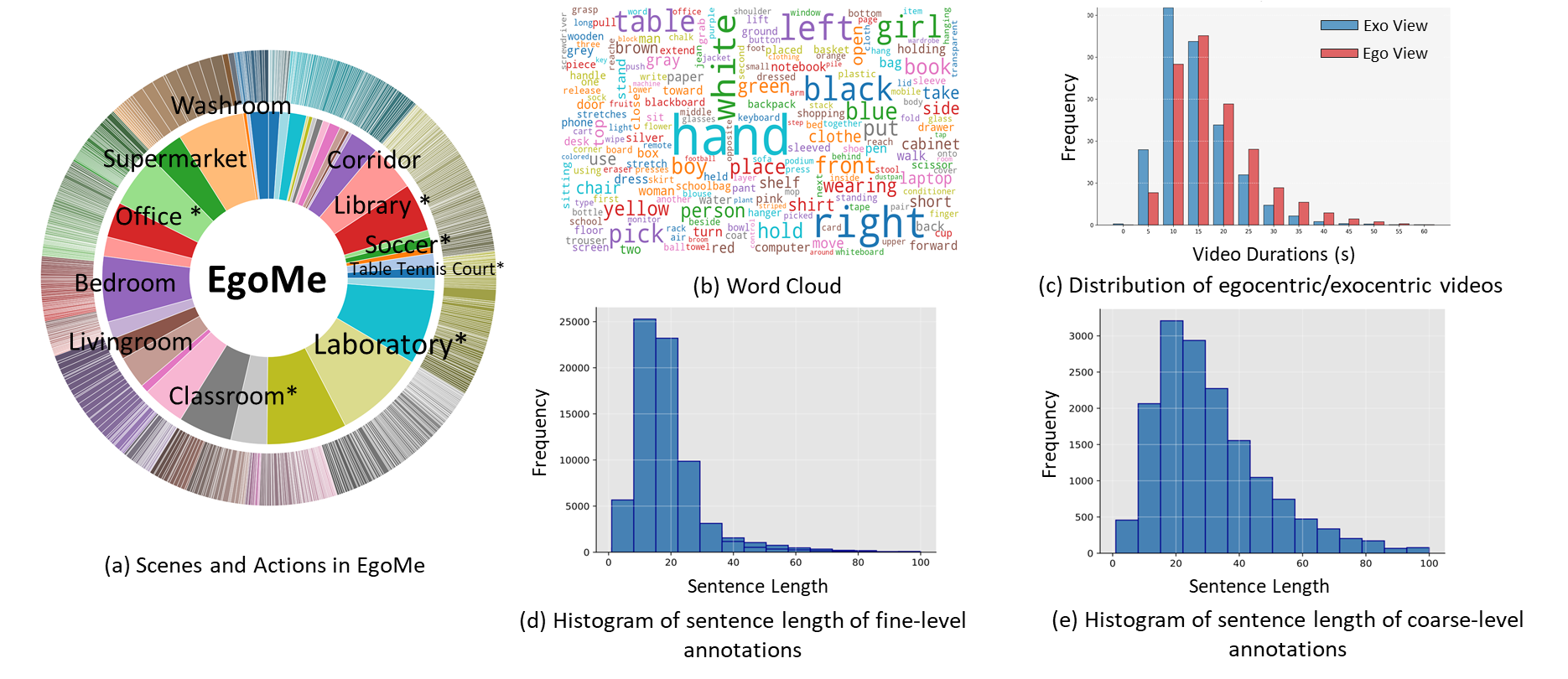}
	\caption{Statistics of the proposed EgoMe dataset.}
	\label{fig:statistic}
\end{figure*}
\noindent\textbf{Coarse-level Language Annotation.} The coarse-level language annotation aims to provide a global description of the entire egocentric or exocentric video using natural language sentences for subsequent video understanding. Compared to previous datasets \cite{li2024egoexo,huang2024egoexolearn}, our annotations not only cover the action description but also include interactive objects information such as location, relationship, and other attributes. Furthermore, we consider different requirements for the two video views: egocentric videos focus more on detailed descriptions of the hand-object actions, while exocentric videos concentrate on additional global information descriptions like actor and environment characteristics.

%
%
%

\noindent\textbf{Fine-level Language Annotation}
Based on coarse-level language annotation, fine-level video language annotation further divides the exocentric or egocentric video into fine-grained action substeps and provide the corresponding detailed sentence descriptions. In this stage, three types of annotations are required: 1) the start timestamps and 2) end timestamps of each action substep; 3) detailed descriptions of the corresponding action within the time interval, similar to the description requirements of coarse-level annotations. Specifically, we request annotators to treat each hand-object interaction as an independent step and provide detailed language descriptions including ``which hand is used to conduct the current action". In addition, if the imitator in the egocentric video correctly mimics the action in the corresponding exocentric video, the number of action timestamps and contents of the exocentric-egocentric video pair should be consistent.


\noindent\textbf{Imitative Verification Annotation.}
To empower AI agents to understand and localize incorrect mimicking actions in egocentric videos, we provide imitative verification annotation. In detail, if the imitator's action in the egocentric video are identical to the corresponding exocentric video, the video pair is labeled``True"; otherwise, it is labeled ``False". For the case of ``False", annotators must marks the specific wrong action steps in the egocentric videos.


%

\subsection{Statistics and Comparisons}
\label{sec:stat}
As shown in Table\ref{dataset_compar}, our EgoMe dataset collects 15804 videos, comprising 7902 exocentric (observing) videos and 7902 egocentric (following) videos, with a total video duration of 82 hours and 46 minutes and covering up to 41 real-world scenarios184 diverse activity categories in \ref{fig:statistic}(a). Compared with existing datasets, our EgoMe dataset is the first dataset focusing on cross-view video research for imitating learning, featuring a larger number of videos, more diverse scenes and action categories, while avoiding the redundancy of existing long videos. Fig. \ref{fig:statistic}(c) show the distributions of exocentric and egocentric video durations, respectively, ranging widely from 3 seconds to 60 seconds, with most lasting 10 to 25 seconds. 

For comparison of annotation richness, our dataset contains more useful multimodal information, such as eye gaze and other sensor IMU data like quaternion, angular velocity, acceleration, and magnetic strength, which are beneficial for more comprehensive research on human eye movement and body motion information in imitation learning. Moreover, unlike existing datasets that share the same language annotations for a pair of exo-ego videos, we separately labeled the coarse-level and fine-level annotations for videos from different views. In total, there are 15804 global coarse-level descriptions and 69880 fine-level descriptions. Due to the varying complexity of behaviors in the videos, there is a difference in the number of fine-grained annotations. 
Furthermore, we analyze the sentence length of the language annotations. The average sentence length is 32.15 words and 18.15 words, respectively as shown in Figure \ref{fig:statistic}(d). For more details about dataset analyze, please refer to supplementary material.

\begin{figure*}[!t]
	\centering
	\includegraphics[width=0.8\linewidth]{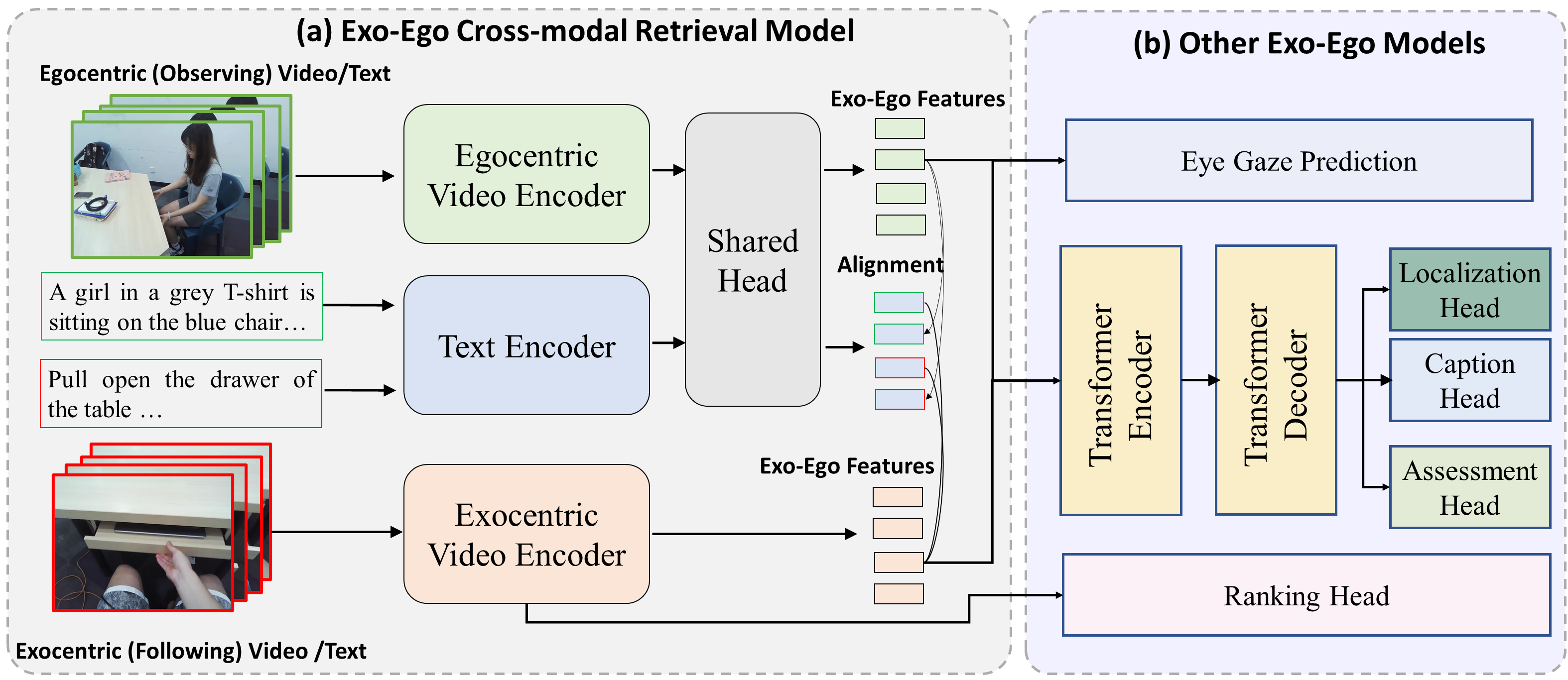}
	\caption{Overview of all baseline models for the proposed benchmarks.}
	\label{fig:benchmark}
\end{figure*}
\section{Benchmarks}
In this section, we introduce a series of challenging benchmarks and their corresponding baselines to evaluate the ability of the imitating learning, including Exo-Ego cross-modal retrieval, Exo-Ego gaze prediction, imitative action assessment or reorganization, Exo-Ego coarse-level or fine-level video understanding. In addition, our dataset encourages more common benchmarks based on the labeled annotations, such as cross-view video generation and cross-view action classification, and scene classification.
Fig.\ref{fig:benchmark} shows the baseline models for these benchmarks, where we pre-trained a Exo-Ego cross-modal retrieval model to extract cross-view aligned features for subsequent tasks. Here, we demonstrate the experiment results on the EgoMe validate set, test set results in supplementary material.
\subsection{Exo-Ego Cross-modal Retrieval}\label{sec-retrieval}

\begin{table}[htbp!]
	\centering
	\scalebox{0.8}{\setlength{\tabcolsep}{3mm}{
			\begin{tabular}{@{}l|ccccccccc@{}}	
				\toprule			
				Setting&R@1&R@5&R@10& MeanR&MedianR\\
				\midrule
				Exo2Ego&17.61&49.56&66.35&12.39&6.0\\
				Ego2Exo&19.36&51.77&68.19&12.10&5.0\\
				\midrule
				Exo2Text&30.25&69.38&83.38&8.42&3.0\\
				Text2Exo&29.88&69.13&83.5&8.65&3.0\\
				Ego2Text&27.88&66.38&83.38&8.42&3.0\\
				Text2Ego&26.88&63.88&79.75&9.83&3.0\\
				\bottomrule	
	\end{tabular}}}
	\caption{Results of Exo-Ego cross-modal retrieval benchmark on various evaluation settings.}
	\label{retrieval}
\end{table}


\noindent\textbf{Motivation.} Unlike existing retrieval tasks \cite{yu2018joint,xue2023learning,huang2024egoexolearn} that focus solely on cross-view or cross-modal retrieval, we propose an Exo-Ego cross-modal retrieval task based on paired videos from exocentric (observation) and egocentric (imitation) views and their corresponding text descriptions. This benchmark aims to simultaneously address the challenges of view and modal gaps, thus requiring the model deeply understand the visual semantic content and view-specific differences and commonalities to establish the mapping between observation and imitation videos. This can provide more generalized feature representations for subsequent cross-view imitating learning research.

\noindent\textbf{Task definition.}
Given a pair of exo-ego videos $\mathcal{V}^i_{p}=\{\mathcal{V}^i_{exo},\mathcal{V}^i_{ego}\}$ and their corresponding language descriptions $\mathcal{T}^i_{p}=\{\mathcal{T}^i_{exo},\mathcal{T}^i_{ego}\}$, the task (i.e., X2Y) is to query the corresponding video or language Y according to the other view or modality X. Here, we consider 6 task settings to evaluate the retrieval performance, including Exo2Ego and Ego2Exo for cross-view retrieval, Exo2Text, Text2Exo, Ego2Text and Text2Ego for cross-modal retrieval.


\noindent\textbf{Baseline and Results.} To achieve cross-view and cross-modal retrieval simultaneously, we adopted a baseline Exo-Ego cross-modal retrieval model inspired by \cite{xu2024retrieval}, which use VideoMAE-L \cite{tong2022videomae} and UniformerV2 \cite{li2023uniformerv2} model pretrained on Ego4D as egocentric video encoder and third-person video as exocentric video encoder, respectively; and a Bert model as the text encoder. Unlike original InfoNCE loss \cite{radford2021learning}, we adopt EgoExoNCE loss with cross-view video-text pairs to optimize the network. Table \ref{retrieval} reports the experimental results of the baseline model on different evaluation settings. It can be observed that cross-view is more difficult compared with cross-modal retrievals.
\subsection{Exo-Ego Gaze Prediction}
\begin{table}[htbp]
	\centering
	\scalebox{0.8}{\setlength{\tabcolsep}{5mm}{
			\begin{tabular}{@{}l|ccc@{}}	
				\toprule			
				Setting&\multirow{2}{*}{F1}&\multirow{2}{*}{Recall}&\multirow{2}{*}{Precision}\\
				(\emph{Train$\rightarrow$Test})\\
				\midrule
               \emph{Single-view}\\
				Exo $\rightarrow$ Exo \cite{lai2022eye}&56.52&65.56&49.65\\
				Ego $\rightarrow$ Ego \cite{lai2022eye}&47.11&60.40&38.62\\
				\midrule
				\emph{Cross-view}\\
				Exo $\rightarrow$ Ego&41.56&55.81&33.11\\
				Ego$\rightarrow$ Exo&47.65&59.61&39.69\\
				Exo*-Ego$\rightarrow$ Ego &42.11&54.29&34.39\\
				Exo-Ego*$\rightarrow$ Exo &55.81&62.40&50.73\\
				\midrule
				\emph{Co-training}\\
				Exo\&Ego $\rightarrow$ Exo&56.48&64.51&50.23\\
				Exo\&Ego $\rightarrow$ Ego&47.32&59.67&39.20\\
				\bottomrule	
	\end{tabular}}}
	\caption{Results of Exo-Ego gaze prediction benchmark on various training and evaluation settings, where X*-Y represents using the X supervised information during training, and the X and Y features extracted from model in Sec.\ref{sec-retrieval}.}
	\label{gaze}
\end{table}
%
%
\noindent\textbf{Motivation.} Eye gaze prediction offers valuable insights into visual attention mechanisms for human cognition process. However, existing research \cite{lai2022eye,lai2023eye} primarily focuses on ego-view gaze prediction, neglecting gaze patterns when human observes the actions of others via exocentric view, which is important for understanding the attention processing about how human imitates others' actions. Thus, we propose the first Exo-Ego gaze prediction task to establish a correlation between gaze attention map of observing and imitating process, thereby deeply decoding cross-view attention transition.


\noindent\textbf{Task definition.} Given a pair of Exo-Ego videos $\mathcal{V}^i_{p}=\{\mathcal{V}^i_{exo},\mathcal{V}^i_{ego}\}$ and eye gaze information  $\mathcal{G}^i_{p}=\{\mathcal{G}^i_{exo},\mathcal{G}^i_{ego}\}$, during training, there only are ground truth of gaze point information from an exocentric observation video,  the model is ask to transfer the gaze information to infer possible gaze point locations in the egocentric imitating video by building the establish the eye gaze correlation between observing (Exo) and following (Ego) videos, and vice versa.

\noindent\textbf{Baseline and Results.} We adopt typical egocentric gaze prediction model GLC \cite{lai2022eye} as single-view baseline.  To achieve cross-view alignment, we further fuse the aligned view invariable Exo-Ego feature representation extracted from Sec. \ref{sec-retrieval} to evaluate the gaze prediction performance. As shown in Table \ref{gaze}, it can be observed that the cross-view aligned features method achieve better results compared with naive cross-view learning, which demonstrates the effectiveness of the proposed cross-view alignment method.

\subsection{Imitative Action Assessment}\label{sec-assessment}
%
\begin{table}[htbp]
	\centering
	\scalebox{0.8}{\setlength{\tabcolsep}{4mm}{
			\begin{tabular}{@{}l|ccccc@{}}	
				\toprule			
				Setting&AP@30&AP@50&AP@70\\
				\midrule
				Exo-Ego (TSP \cite{alwassel2021tsp})&39.31&28.87&22.05\\		
				\midrule
				Exo-Ego (Our) &42.74&32.61&22.63\\
				\bottomrule	
	\end{tabular}}}
	\caption{Results of imitative action assessment benchmark.}
	\label{assessment}
\end{table}

\noindent\textbf{Motivation.} Imitative action assessment is desirable to study advancing embodied imitation learning, requiring fine-grained analysis of action step consistency across cross-view observing and following video pairs (exo-ego). While there are some analogous benchmarks (e.g., single-view error detection \cite{lee2024error} and sequential verification \cite{qian2022svip}), they fail to address the compounded challenges of the proposed cross-view imitative action assessment. In contrast, we introduce a new and more challenge paradigm, which not only requires the model to locate the fine-grained action steps, but also identify whether the step is correct or error within the imitated video. 

%
%
%
%

\noindent\textbf{Task definition.} 
Given a pair of observing and following videos $\mathcal{V}^i_{p}=\{\mathcal{V}^i_{exo},\mathcal{V}^i_{ego}\}$, each video is divided into multiple steps, which are represented by a series of time periods, i.e., $\{(t^s_j,t^e_j, M_j)\}_{j=1}^{N_{step}}$, where $t^s_j$, $t^e_j$, $M_j$ and $N_{step}$ are the starting time, ending time and the number of action steps in each egocentric video. $M_j=\{0,1\}$, the $j$-th step is imitated correctly in the following (ego) video when $M_j=1$, and vice versa. This task asks to predict the time location of each action step and imitative action score according to the exocentric videos.


\noindent\textbf{Baseline and Results.} We first extract the exocentric and egocentric video features, and then introduce a localization head and a binary classification head based on the Deformable Transformer to predict the event localization and action score. Similar to eye gaze prediction, we also replace video feature using our cross-view aligned feature representation. As shown in Table\ref{assessment}, it can be observed that our method outperforms the Exo-Ego features extracted from single-view TSP encoder \cite{alwassel2021tsp}.

%

\subsection{Imitative Action Reorganization}
			%
\begin{table}[htbp]
	\centering
	\scalebox{0.8}{\setlength{\tabcolsep}{5mm}{
			\begin{tabular}{@{}l|ccccc@{}}	
				\toprule			
				Setting&Ranking Accuracy\\
				\midrule
				Exo*-Ego$\rightarrow$ Exo &18.61\\		
				Exo-Ego*$\rightarrow$ Ego &21.25\\
				\bottomrule	
	\end{tabular}}}
\caption{Results of imitate action reorganization benchmark.}
\label{reorganization}
\end{table}

\noindent\textbf{Motivation.} In addition to the above imitative action assessment benchmark, the imitate action reorganization remains important to verify the imitation learning ability of the model, but it is an unexplored aspect in current video understanding research. Our work pioneers a cross-view imitate action reorganization benchmark, which not only enhances the robot's ability to acquire and imitate human actions, but more importantly, by establishing the interaction relationships between some simple action sub-steps, it enables the effective understanding of more long and complex action tasks in the future.

\noindent\textbf{Task definition.} Similar to Sec.\ref{sec-assessment}, each video is segmented into fine-grained action fragments i.e., $\{(t^s_j,t^e_j)\}_{j=1}^{N_{step}}$ that capture simple action sub-steps within a activity. The model is tasked with inferring the correct order of these fragments, enabling the reconstruction of the overall egocentric video, which can reflect the natural progression of the complete action sequence, based on the temporal and contextual relationships between them. 
%

\noindent\textbf{Baseline and Results.} To achieve this task, we feed the overall exocentric video and a set of shuffled video fragments from its corresponding egocentric video into the cross-view retrieval model, and adopt EgoExoNCE loss with cross-view video-text pairs to optimize the network. Table\ref{reorganization} shows the performance of ranking videos from another view based on either an ego view or an exo view video.

\subsection{Exo-Ego Coarse-level Video Understanding}
\begin{table}[htbp]
	\centering
	\scalebox{0.8}{\setlength{\tabcolsep}{2mm}{
			\begin{tabular}{@{}l|ccccccccc@{}}	
				\toprule			
			  Setting&\multirow{2}{*}{BLEU@4}&\multirow{2}{*}{METEOR}&\multirow{2}{*}{ROUGE-L}&\multirow{2}{*}{CIDER}\\
				(\emph{Train$\rightarrow$Test})\\
				\midrule
				\emph{Single-view}\\
				Exo $\rightarrow$ Exo \cite{wang2021end}&8.21&13.95&33.53&32.37\\
				Ego $\rightarrow$ Ego \cite{wang2021end}&7.13&13.03&33.67&43.65\\
				\midrule
				\emph{Cross-view}\\
				Exo $\rightarrow$ Ego&6.7&12.19&27.05&23.96\\
				
				Ego$\rightarrow$ Exo&3.45&9.37&24.89&14.40\\
Exo-Ego*$\rightarrow$ Ego &8.25&13.75&34.80&45.26\\
				Exo-Ego*$\rightarrow$ Exo &2.70&10.03&20.27&3.02\\
				\midrule
				\emph{Co-training}\\
				Exo\&Ego $\rightarrow$ Exo &8.43&14.13&34.08&34.55\\
				Exo\&Ego $\rightarrow$ Ego&9.12&14.53&35.04&49.02\\
				\bottomrule	
	\end{tabular}}}
	\caption{Results of coarse-level Exo-Ego video understanding benchmark on various training and evaluation settings.}
	\label{coarse}
\end{table}
\noindent\textbf{Motivation.}Video understanding is a hot topic that aims to parse the visual scenes and generate textual descriptions for the given video. Most prior works \cite{wang2021pos,tang2021clip4caption,dai2024egocap,nakamura2021sensor} perform video captioning of single-view videos with remarkable success. However, cross-view correlation and joint understanding are essential since egocentric videos provide local details of human activities, while exocentric videos contain more global visual contextual information. Therefore, we propose an Exo-Ego coarse-level video understanding benchmark.

\noindent\textbf{Task definition.} In our EgoMe dataset, there are the paired exocentric and egocentric videos  $\mathcal{V}^i_{p}=\{\mathcal{V}^i_{exo},\mathcal{V}^i_{ego}\}$ with their corresponding coarse-level global textual description annotations $\mathcal{C}^i_{p}=\{\mathcal{C}^i_{exo},\mathcal{C}^i_{ego}\}$. During training, the paired videos with their corresponding coarse-level global textual description annotations are given to perform the cross-view joint training to the model's learning of exo-ego complementary information. During inference, the model is tested on exocentric or egocentric videos respectively to generate the corresponding textual video captions. 

\noindent\textbf{Baseline and Results.} To achieve this Exo-Ego coarse-level video understanding task, we adopt a video caption head with soft-attention enhanced LSTM following \cite{wang2021end} as our baseline model. In Table\ref{coarse}, it was an unexpected finding that we used aligned features to evaluate on cross-view Exo videos, but the performance decreased in some indicators compared to the naive cross-view method, it may be Exo videos often include information such as the demonstrator's clothes, gender, etc., which is almost non-existent in ego videos, making it difficult to achieve precise language descriptions.
%
%

\subsection{Exo-Ego Fine-level Procedural Understanding}
\begin{table}[htbp]
	\centering
	\scalebox{0.8}{\setlength{\tabcolsep}{0.8mm}{
		\begin{tabular}{@{}l|ccc|ccc@{}}	
		\toprule			
  Setting&\multicolumn{3}{c|}{dvc\_eval}&\multicolumn{3}{c}{SODA}\\
       \emph{Train$\rightarrow$Test}&BLEU@4&METEOR&CIDER&METEOR&CIDER&tIoU\\
		\midrule
		\emph{Single-view}\\
		Exo $\rightarrow$ Exo \cite{wang2021end}&4.53&12.08&33.01&10.60&27.06&50.41\\
		Ego $\rightarrow$ Ego \cite{wang2021end}&5.97&13.13&53.01&12.5&42.03&49.86\\
		\midrule
		\emph{Cross-view}\\
		Exo $\rightarrow$ Ego&2.53&10.51&24.22&7.57&19.22&47.76\\
		Ego$\rightarrow$ Exo&1.63&8.02&16.40&8.60&15.93&47.89\\
		Exo-Ego*$\rightarrow$ Exo &1.54&7.56&15.88&8.42&11.81&46.27\\
		Exo-Ego*$\rightarrow$ Ego &6.15&13.41&55.03&13.39&45.79&51.12\\
		\midrule
		\emph{Co-training}\\
		Exo\&Ego $\rightarrow$ Exo &5.06&12.65&36.09&10.88&29.41&49.55\\
		Exo\&Ego $\rightarrow$ Ego &6.41&13.46&56.63&13.08&46.13&51.02\\
		\bottomrule	
\end{tabular}}}
	\caption{Results of Exo-Ego fine-level video understanding benchmark on various training and evaluation settings.}
	\label{fine}
\end{table}
\noindent\textbf{Motivation.} With the development of embodied AI, the comprehensive and more fine-level procedural understanding of human activities in videos has become increasingly essential for intelligent agents. However, previous researchers \cite{chen2023egoplan,qiu2024egoplan,cheng2024videgothink} primarily focused on single-view fine-grained video understanding, a recent research has begun to explore unpaired fine-grained understanding, due to the lack of such datasets that simultaneously possess paired Exo-Ego fine-level language annotations. Instead of them, our EgoMe dataset consists of paired Exo-Ego videos with fine-level descriptions for each sub-step, which is beneficial in enabling models to understand long videos in greater detail, model the relationships between each step, and transfer current extensive exocentric video knowledge to egocentric video understanding.


\noindent\textbf{Task definition.} Given a pair of videos $\mathcal{V}^i_{p}=\{\mathcal{V}^i_{exo},\mathcal{V}^i_{ego}\}$ with their corresponding fine-level textual description annotations $Y_{exo/ego}\{(t^s_j,t^e_j, F_j)\}_{j=1}^{N_{step}}$ where $F_j$ is the fine-level caption for $j$-th timestamp in the exocentric or egocentric video. Compared with coarse-level understanding, this task aims to anticipate the fine-grained description and timestamp based on the given exo-ego videos. 

\noindent\textbf{Baseline and Results.} We adopt PDVC \cite{wang2021end} as our sing-view fine-level video caption baseline model. In addition, we also perform the cross-view alignment (Ego-Exo) for fine-level video understanding. Here is a similar finding in Table\ref{fine}, where the cross-view alignment features is slightly lower than the naive cross-view method on exocentric videos, which may make it more difficult to capture the demonstrator's appearance information directly from an egocentric video.

\section{Conclusion}
In this paper, we introduce a  novel egocentric dataset, EgoMe, which aims to follow human behavior via egocentric view in the real world. Unlike existing researches, our dataset takes the egocentric view as a benchmark to record 7902 pairs of observing and following videos (15804 videos) for diverse daily behaviors in real-world scenarios. It can better reflect the process of human learning from observation to imitation. The dataset also contains eye gaze and other sensor multimodal data for assisting in establishing correlations between observation and imitation behaviors. In addition, we also propose several challenging benchmark tasks for fully leveraging this data resource and corresponding baseline models for promoting the research of robot imitation learning ability. We hope that our EgoMe dataset can contribute significantly to the future development of embodied intelligence and AR/VR. 
{
    \small
    \bibliographystyle{ieeenat_fullname}
    \bibliography{main}
}

\maketitlesupplementary
\setcounter{section}{5}
In this supplementary material, we provide additional details about the collection and annotation of the dataset, and additional our benchmark details.
\section{Dataset Details}
In our EgoMe dataset, we consider daily human activities and their follow-up learning in real-world environments, therefore we collect data in up to over 41 scenarios including various classrooms, living rooms, supermarkets, offices, libraries, washrooms, gardens, gymnasiums, and so on. In addition, each kind of scenario comprises a diverse range of layouts. These activities include simple activities such as typing words on the phone and opening the door, and complex activities such as binding documents, folding clothes, and using various instruments. In addition, considering the potential applications of scenes where AI robots assist or nurse humans, we innovatively introduce assisting and nursing-related activities, e.g. lifting objects with humans, helping humans wear jewelry or clothes, assisting humans wash up, and so on. Moreover, when recording videos with a particular activity category, different videos also contain distinct subjects, actors, scenarios, camera angles, and contextual actions to guarantee the diversity of the proposed dataset. 

\begin{figure}[htbp!]
	\centering
	\begin{subfigure}{0.8\linewidth}
		\centering
		\includegraphics[width=1.0\linewidth]{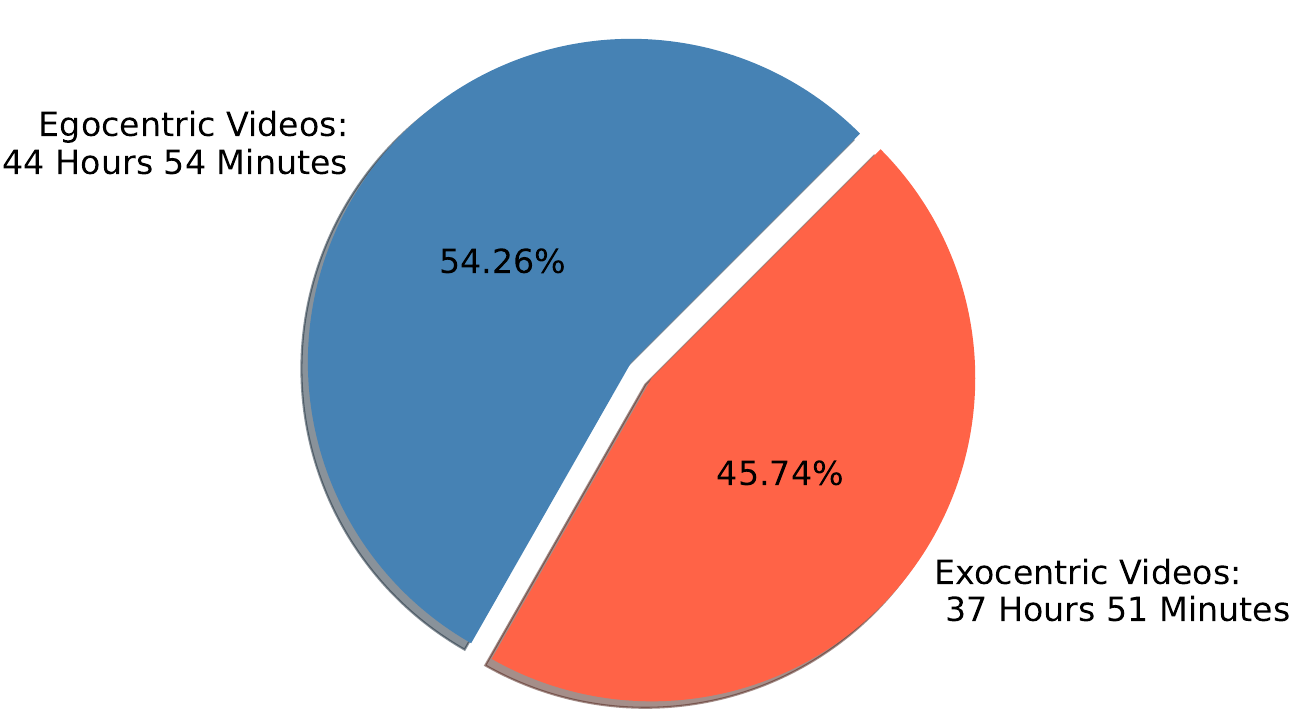}
		\caption{The ratio of exocentric and egocentric video duration.}
		\label{fig:pie_chart1}
	\end{subfigure}
	\begin{subfigure}{0.8\linewidth}
		\centering
		\includegraphics[width=1.0\linewidth]{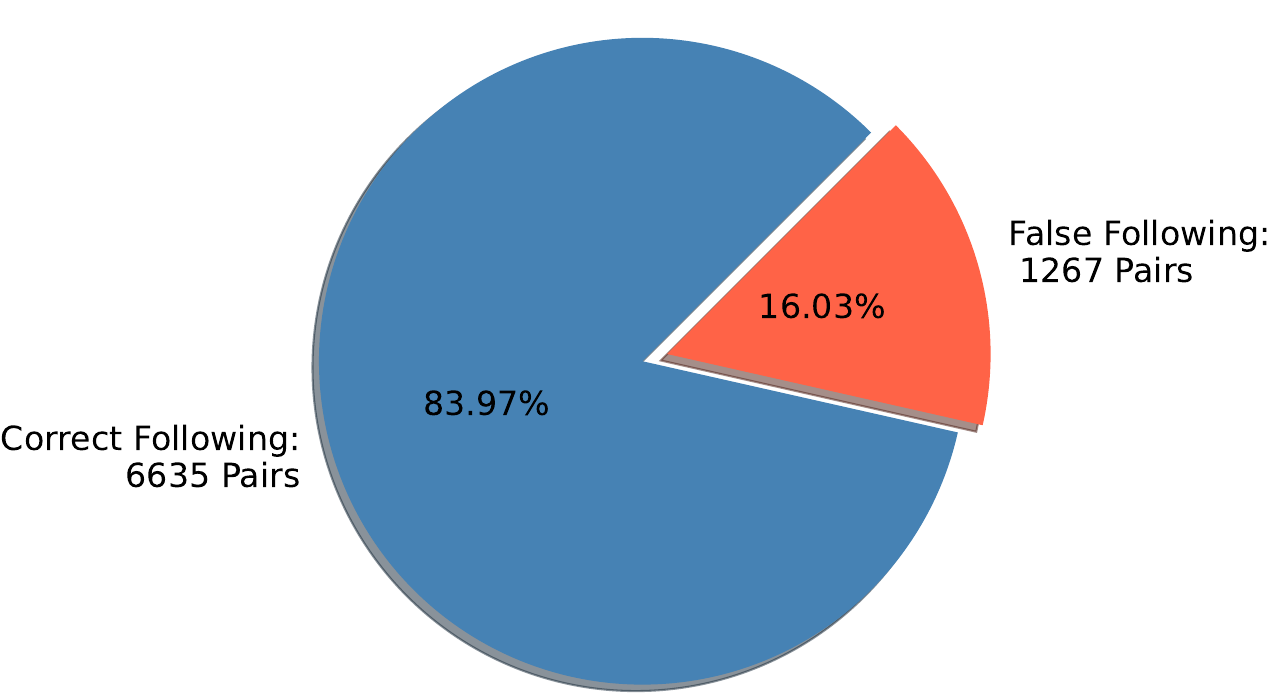}
		\caption{The ratio of correct and false following video pairs.}
		\label{fig:pie_chart2}
	\end{subfigure}
	\hfill
	\caption{Statistic of exocentric/egocentric video duration and correct/incorrect following video pairs.}
	\label{fig:pie_chart}
\end{figure}

\begin{figure*}[htbp]
	\centering
	\includegraphics[width=0.9\linewidth]{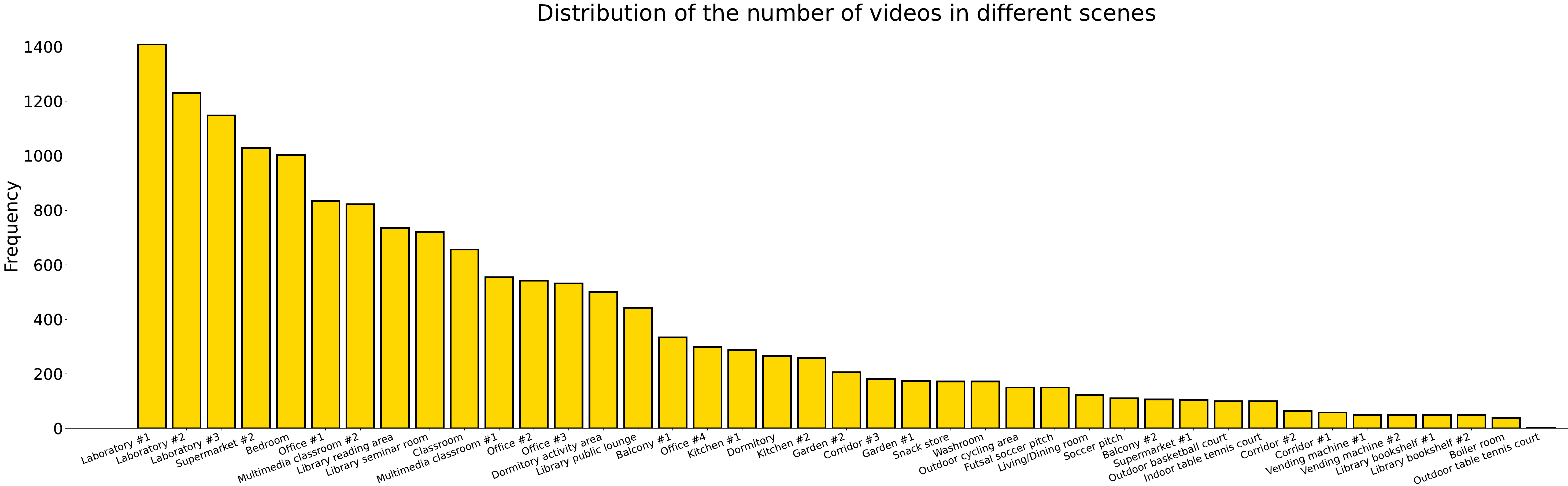}
	\caption{Distribution of occurrences of videos in the EgoMe dataset in different scenarios.}
	\label{fig:scene_bar}
\end{figure*}

\begin{figure*}[ht]
	\centering
	\begin{subfigure}[htbp]{1.0\textwidth}
		\includegraphics[width=1.0\linewidth]{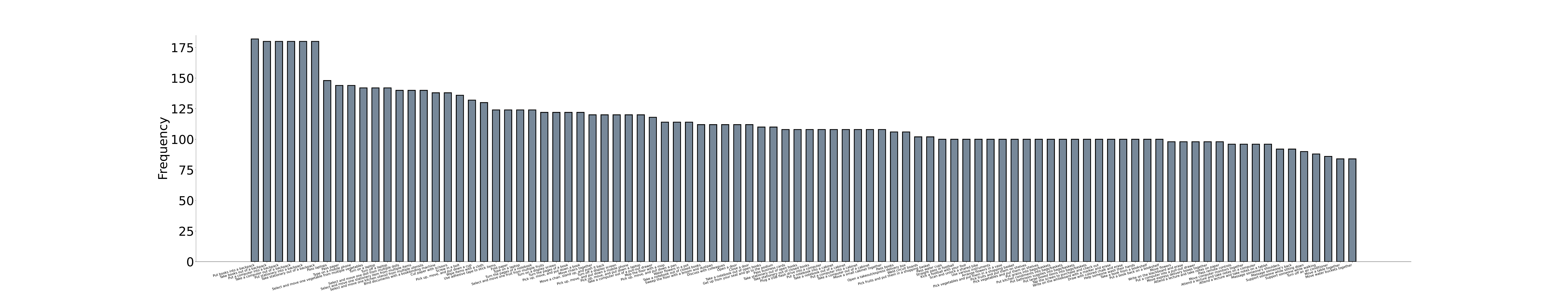}
		\caption{The bar chart of the number of videos in different activities (the front part).}
		\label{fig:class_bar_1}
	\end{subfigure}
	\begin{subfigure}[htbp]{1.0\textwidth}
		\includegraphics[width=1.0\linewidth]{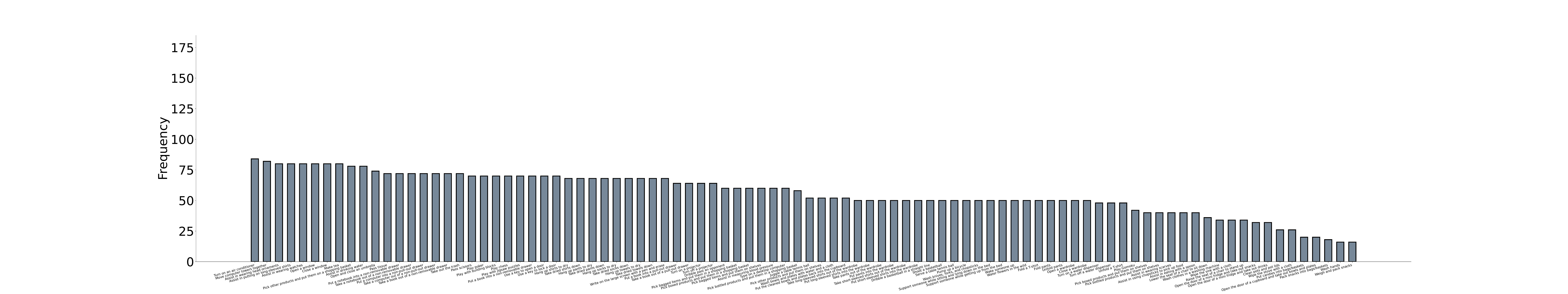}
		\caption{The bar chart of the number of videos in different activities (the latter part).}
		\label{fig:class_bar_2}
	\end{subfigure}
	\caption{Occurrence distribution of videos in the EgoMe dataset in different activity categories.}
	\label{fig:class_bar}
\end{figure*}

As shown in Fig.\ref{fig:scene_bar} and Table \ref{tab:scene_categories}, there are up to 41 real-world scenarios in the proposed dataset such as indoor offices, classrooms, supermarkets, kitchens, and outdoor gardens, basketball courts, cycling areas. Fig.\ref{fig:class_bar} presents the occurrence distribution of exocentric and egocentric videos in different activity categories. Fig. \ref{fig:pie_chart1} demonstrates that the proportion of video durations for egocentric and egocentric videos is almost evenly split, which shows that the total duration of exocentric videos is 37 hours and 51 minutes, which occupies 45.74$\%$ of the total duration of the dataset. The total duration of egocentric videos is 44 hours and 54 minutes, accounting for 54.26$\%$. 

\begin{figure*}[htbp]
	\centering
	\includegraphics[width=1.0\linewidth]{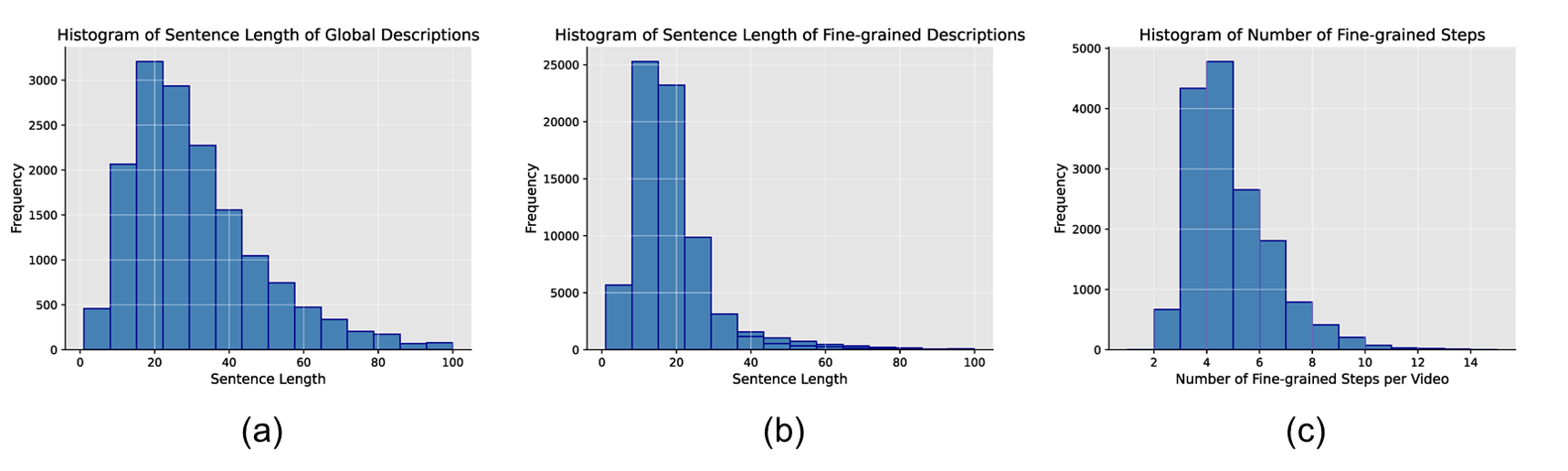}
	\caption{Distribution of sentence length of global and fine-grained descriptions.}
	\label{fig:sentence}
\end{figure*}

\begin{figure*}[htbp!]
	\centering
	\includegraphics[width=1.0\linewidth]{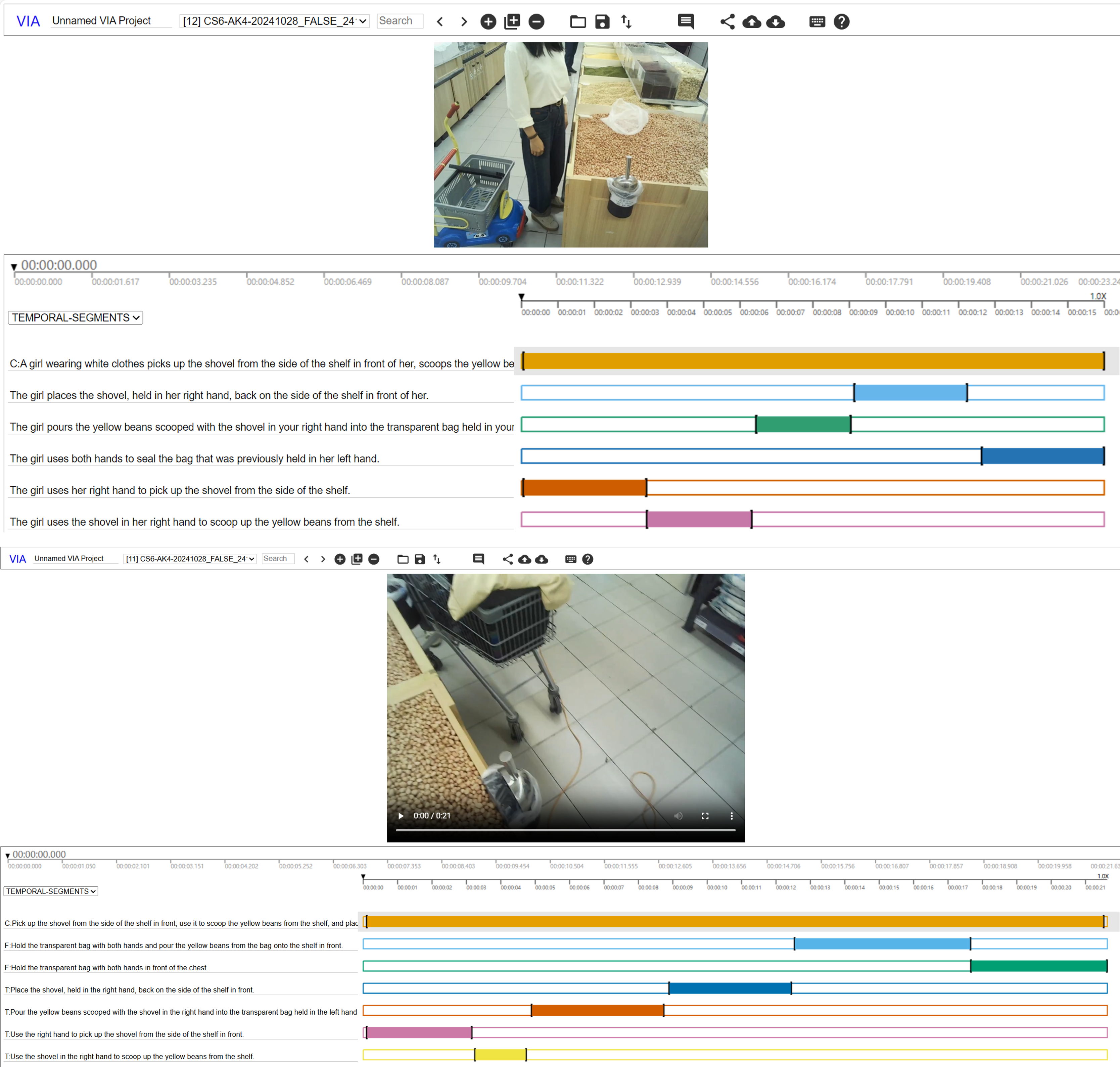}
	\caption{The annotation software interfaces for our EgoMe dataset.}
	\label{fig:sup_label}
\end{figure*}

\begin{figure*}[htbp!]
	\centering
	\includegraphics[width=1.0\linewidth]{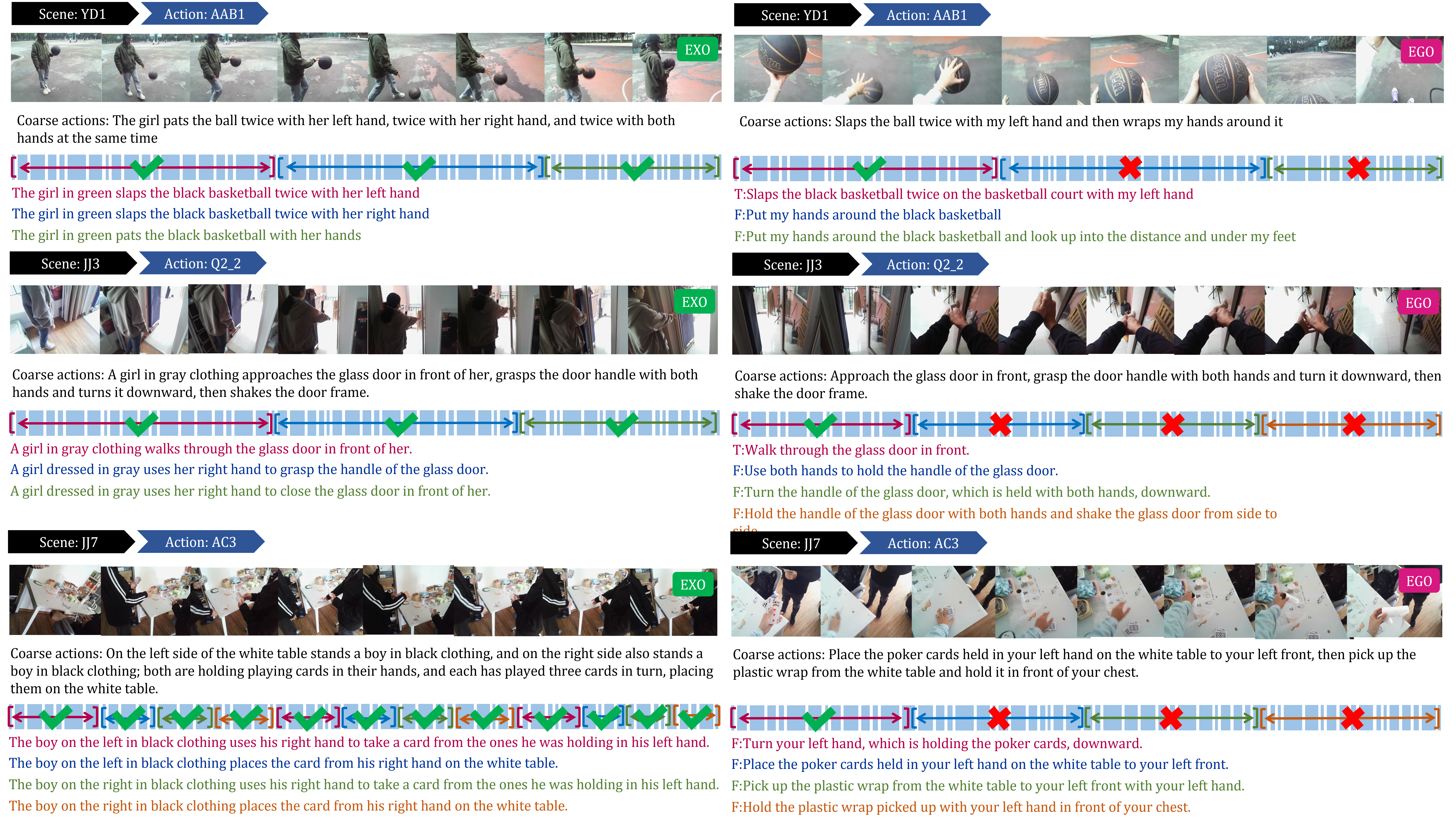}
	\caption{Illustration of more examples in our EgoMe dataset.}
	\label{fig:sup_show}
\end{figure*}

In addition, each video is annotated with a global description and multiple fine-grained descriptions. In total, we have annotated 15804 global descriptions and 69880 fine-grained descriptions.  Due to the varying complexity of behaviors in the videos, there is a difference in the number of fine-grained annotations. As shown in Figure \ref{fig:sentence}, the maximum number of fine-grained descriptions for a single video is 15, while the minimum is 2. Additionally, we have counted the number of words in the global descriptions and fine-grained descriptions, which totals 1776040 words. Fig\ref{fig:sup_label} and Fig\ref{fig:sup_show} shows the annotation software interfaces for our EgoMe dataset, and more visualization examples for our EgoMe dataset.
\begin{table*}[h]
	\centering
	\caption{Category and Scene Correspondence.}
	\label{tab:scene_categories}
	\begin{tabular}{ll}
		\toprule
		\textbf{Category} & \textbf{Scenes} \\
		\midrule
		Classroom (JS) & JS1. Pinxue Building, Liren Building A and B Classrooms \\
		& JS2. Pinxue Building C Classrooms \\
		& JS3. Pinxue Building Corridor \\
		& JS4. Liren Building Corridor \\
		& JS5. Business Management Building Classrooms \\
		\midrule
		Conference Room (HYS) & HYS1. KB Fourth Floor Conference Room \\
		& HYS2. KB302 Conference Room (Available for Booking) \\
		& HYS3. Other Bookable Conference Rooms \\
		\midrule
		Kitchen (CF) & CF1. Qiu's Kitchen \\
		& CF2. Xiao's Kitchen \\
		& CF3. He’s Small Kitchen \\
		& CF4. 535's Simple Kitchen \\
		\midrule
		Supermarket (CS) & CS1. Campus Supermarket \\
		& CS2. Hongqi Supermarket \\
		& CS3. Vending Machines \\
		& CS4. EasyBuy \\
		& CS5. Commercial Street Second Floor Grocery Store \\
		& CS6. Yonghui Supermarket \\
		\midrule
		Sports (YD) & YD1. Outdoor Basketball Court \\
		& YD2. Indoor Basketball Court \\
		& YD3. Indoor Badminton Hall \\
		& YD4. Outdoor Badminton Court \\
		& YD5. Indoor Table Tennis Area \\
		& YD6. Outdoor Table Tennis Area \\
		& YD7. Outdoor Tennis Court \\
		& YD8. Small Soccer Field \\
		& YD9. Large Soccer Field \\
		\midrule
		Reading Area (YDL) & YDL1. Library Reading Room \\
		& YDL2. First Floor Library Party-Building Bookshelf \\
		& YDL3. General Library Bookshelf \\
		& YDL4. Library PinYue Self-Study Room \\
		\bottomrule
	\end{tabular}
\end{table*}


\section{Benchmark Details}
Similar to benchmark section in formal content paper, we also conduct all experiments on the test set of our EgoMe dataset, including Exo-Ego cross-modal retrieval, Exo-Ego gaze prediction, imitative action assessment or reorganization, Exo-Ego coarse-level or fine-level video understanding. A shown in Table \ref{retrieval,gaze,assessment,reorganization,reorganization,coarse,fine}, it can be observed that these results on the test set are a similar experimental results to validate set in our EgoMe dataset.
\begin{table}[htbp!]
	\centering
	\scalebox{1.0}{\setlength{\tabcolsep}{3mm}{
			\begin{tabular}{@{}l|ccccccccc@{}}	
				\toprule			
				Setting&R@1&R@5&R@10& MeanR&MedianR\\
				\midrule
				Exo2Ego&17.61&49.56&66.35&12.39&6.0\\
				Ego2Exo&19.36&51.77&68.19&12.10&5.0\\
				\midrule
				Exo2Text&18.87&51.73&70.54&15.39&5.0\\
				Text2Exo&19.35&51.13&69.81&15.42&5.0\\
				Ego2Text&16.76&48.94&66.91&17.18&6.0\\
				Text2Ego&15.76&46.56&65.37&17.38&6.0\\
				\bottomrule	
	\end{tabular}}}
	\caption{Results of Exo-Ego cross-modal retrieval benchmark on various evaluation settings on the test set.}
	\label{retrieval}
\end{table}

\begin{table}[htbp!]
	\centering
	\scalebox{0.8}{\setlength{\tabcolsep}{5mm}{
			\begin{tabular}{@{}l|ccc@{}}	
				\toprule			
				Setting&\multirow{2}{*}{F1}&\multirow{2}{*}{Recall}&\multirow{2}{*}{Precision}\\
				(\emph{Train$\rightarrow$Test})\\
				\midrule
				\emph{Single-view}\\
				Exo $\rightarrow$ Exo &56.87&66.09&49.91\\
				Ego $\rightarrow$ Ego &47.66&61.12&39.06\\
				\midrule
				\emph{Cross-view}\\
				Exo $\rightarrow$ Ego&42.00&56.72&33.35\\
				Ego$\rightarrow$ Exo&48.26&60.49&40.15\\
				Exo*-Ego$\rightarrow$ Ego &42.47&55.10&34.55\\
				Exo-Ego*$\rightarrow$ Exo &56.27&62.95&50.86\\
				\midrule
				\emph{Co-training}\\
				Exo\&Ego $\rightarrow$ Exo&57.02&65.18&50.67\\
				Exo\&Ego $\rightarrow$ Ego&47.90&60.20&39.77\\
				\bottomrule	
	\end{tabular}}}
	\caption{Results of Exo-Ego gaze prediction benchmark on various training and evaluation settings on the test set.}
	\label{gaze}
\end{table}

			%
\begin{table}[htbp!]
	\centering
	\scalebox{0.8}{\setlength{\tabcolsep}{1mm}{
			\begin{tabular}{@{}l|ccccc@{}}	
				\toprule			
				Setting&AP@30&AP@50&AP@70\\
				\midrule
				Exo-Ego (TSP)&38.50&28.87&22.45\\		
				\midrule
				Exo-Ego (Alignment)&41.98&31.78&22.17\\
				\bottomrule	
	\end{tabular}}}
	\caption{Results of imitative action assessment benchmark on various settings on the test set.}
	\label{assessment}
\end{table}

\begin{table}[htbp!]
	\centering
	\scalebox{0.8}{\setlength{\tabcolsep}{3mm}{
			\begin{tabular}{@{}l|ccccc@{}}	
				\toprule			
				Setting&Ranking Accuracy\\
				\midrule
				Exo*-Ego$\rightarrow$ Exo &18.56\\		
				Exo-Ego*$\rightarrow$ Ego &22.28\\
				\bottomrule	
	\end{tabular}}}
	\caption{Results of imitate action reorganization benchmark on various settings on the test set.}
	\label{reorganization}
\end{table}

\begin{table}[htbp!]
	\centering
	\scalebox{0.8}{\setlength{\tabcolsep}{1mm}{
			\begin{tabular}{@{}l|ccccccccc@{}}	
				\toprule			
				Setting&\multirow{2}{*}{BLEU@4}&\multirow{2}{*}{METEOR}&\multirow{2}{*}{ROUGE-L}&\multirow{2}{*}{CIDER}\\
				(\emph{Train$\rightarrow$Test})\\
				\midrule
				\emph{Single-view}\\
				Exo $\rightarrow$ Exo&8.00&14.12&33.46&31.77\\
				Ego $\rightarrow$ Ego &7.00&13.15&33.48&43.78\\
				\midrule
				\emph{Cross-view}\\
				Exo $\rightarrow$ Ego&6.20&12.25&26.75&24.16\\
				Ego$\rightarrow$ Exo&3.39&9.60&25.08&15.54\\
				Exo-Ego*$\rightarrow$ Exo &2.40&9.90&19.72&2.53\\
				Exo-Ego*$\rightarrow$ Ego &8.02&13.96&34.34&46.13\\
				\midrule
				\emph{Co-training}\\
				Exo\&Ego $\rightarrow$ Exo &8.65&14.49&34.29&35.32\\
				Exo\&Ego $\rightarrow$ Ego&8.73&14.55&34.59&51.83\\
				\bottomrule	
	\end{tabular}}}
	\caption{Results of coarse-level exo-ego video understanding benchmark on various training and evaluation settings on the test set.}
	\label{coarse}
\end{table}

\begin{table}[htbp!]
	\centering
	\scalebox{0.8}{\setlength{\tabcolsep}{0.8mm}{
			\begin{tabular}{@{}l|ccc|ccc@{}}	
				\toprule			
				Setting&\multicolumn{3}{c|}{dvc\_eval}&\multicolumn{3}{c}{SODA}\\
				\emph{Train$\rightarrow$Test}&BLEU@4&METEOR&CIDER&METEOR&CIDER&tIoU\\
				\midrule
				\emph{Single-view}\\
				Exo $\rightarrow$ Exo &3.40&11.74&30.85&10.55&26.06&50.63\\
				Ego $\rightarrow$ Ego &5.62&12.76&49.01&12.42&41.18&49.62\\
				\midrule
				\emph{Cross-view}\\
				Exo $\rightarrow$ Ego&2.63&10.48&24.45&7.71&18.42&48.81\\
				Ego$\rightarrow$ Exo&1.60&7.91&15.94&8.56&14.39&48.13\\
				Exo-Ego*$\rightarrow$ Ego &6.29&13.22&53.90&13.35&44.46&51.02\\
				Exo-Ego*$\rightarrow$ Exo &1.36&7.52&14.43&8.60&10.67&47.26\\
				\midrule
				\emph{Co-training}\\
				Exo\&Ego $\rightarrow$ Exo&4.43&12.30&33.67&10.69&27.90&49.42\\
				Exo\&Ego $\rightarrow$ Ego &6.22&13.26&53.38&13.02&43.25&50.96\\
				\bottomrule	
	\end{tabular}}}
	\caption{Results of Exo-Ego fine-level video understanding benchmark on various training and evaluation settings on the test set.}
	\label{fine}
\end{table}

\end{document}